\documentclass[a4paper]{article}
\usepackage{spconf,amsmath,graphicx}
\usepackage{amsfonts}
\usepackage{amssymb}
\usepackage{multirow}
\usepackage{bm}
\usepackage{subfigure}
\usepackage{url}
\usepackage{multirow}
\usepackage{algorithm}
\usepackage{algorithmicx}
\usepackage{algpseudocode}
\usepackage{lipsum}
\usepackage{multicol}

\title{Directional Regularized Tensor Modeling for Video Rain Streaks Removal}
\name{Zhaoyang Sun$^{\dag}$, Shengwu Xiong$^{\dag}$, and Ryan Wen Liu$^{\dag,\ddag,\star}$\thanks{This work was supported by NSFC (No.: 51609195), and Fund of Hubei Key Lab. of Transportation Internet of Things (No.: WHUTIOT-2017B003).}}
\address{$^{\dag}$Hubei Key Lab. of Transportation Internet of Things, School of Computer Science and Technology,\\$^{\ddag}$School of Navigation, Wuhan University of Technology, Wuhan, China\\$^{\star}$Email: wenliu@whut.edu.cn}
\begin{document}
%
\maketitle
\begin{abstract}
Outdoor videos sometimes contain unexpected rain streaks due to the rainy weather, which bring negative effects on subsequent computer vision applications, e.g., video surveillance, object recognition and tracking, etc. In this paper, we propose a directional regularized tensor-based video deraining model by taking into consideration the arbitrary direction of rain streaks. In particular, the sparsity of rain streaks in spatial and derivative domains, the spatiotemporal sparsity and low-rank property of video background are incorporated into the proposed method. Different from many previous methods under the assumption of vertically falling rain streaks, we consider a more realistic assumption that all the rain streaks in a video fall in an approximately similar arbitrary direction. The resulting complicated optimization problem will be effectively solved through an alternating direction method. Comprehensive experiments on both synthetic and realistic datasets have demonstrated the superiority of the proposed deraining method.
\end{abstract}
\section{Introduction}
\label{sec:intro}
The existence of undesirable rain streaks in images or videos has a significant negative impact on outdoor visual system, such as object detection \cite{de1,de2}, event detection \cite{event}, object recognition \cite{recognition} and tracking \cite{track}. Rain streaks removal from videos or images is thus an important issue and has been investigated extensively in computer vision research.

Inspired by the pioneering deraining research \cite{Garg}, a large amount of rain streaks removal methods have been proposed to enhance imaging performance. The current methods can be divided into two categories: single-image-based methods and video-based methods. For single-image rain removal, Kang \textit{et al.} \cite{Kang} decomposed an input hazy image into low- and high-frequency components, and proposed to separate the rain streaks from the high-frequency component using sparse coding. Luo \emph{et al.} \cite{Luo} removed rain streaks from a single image via highly discriminative sparse coding. Li \textit{et al.} \cite{Li} utilized patch-based Gaussian mixture model (GMM) prior to restore a rain-free image from its rainy version. The restored images, however, often suffer from loss of fine details. For video rain streaks removal, Garg \textit{et al.} \cite{Garg} developed a comprehensive rain detection method based on photometric and dynamic models related to rain streaks. Kim \emph{et al.} \cite{Kim} first estimated an initial rain map using temporal correlation, and then restored the rain-free video by low-rank matrix completion. Jiang \emph{et al.} \cite{Jiang} presented a tensor-based video rain streaks removal model by considering the discriminatively intrinsic characteristics of video background and rain streaks. Wei \textit{et al.} \cite{Wei} proposed to improve deraining results by integrating with the spatiotemporal smoothness of moving objects and low-rank structure of video background. Please refer to Ref. \cite{review} for a survey work on rain streaks removal.

More recently, effective deep learning architectures \cite{Yang,Fu,Liu,LiuYangTIP2019,LiuJiangICASSP2018} have achieved impressive results for both single-image and video deraining. It should be pointed out that the deraining results are strongly depended on the volume and diversity of training datasets. To make deraining more flexible, mainly inspired by the work of \cite{Jiang}, we will propose a directional regularized tensor-based video rain streaks removal model based on the assumption of arbitrary direction of rain streaks. Its satisfactory deraining performance benefits from the directional regularizer and tensor modeling.
\section{Problem Formulation}
\label{sec:problem}
The rainy video $ \mathcal{O} \in \mathbb{R}^{m\times n\times t}$  can be modeled as a linear superimposition of the desired unknown video background $\mathcal{B} \in \mathbb{R}^{m\times n\times t}$ and the rain streaks $ \mathcal{R} \in \mathbb{R}^{m\times n\times t}$, i.e.,
\begin{equation}
   \mathcal{O} = \mathcal{B} + \mathcal{R}.
\end{equation}

The goal of deraining is to decompose the desired video background $\mathcal{B}$ and rain streaks $\mathcal{R}$ from the input rainy video $\mathcal{O}$. This decomposition process is an ill-posed inverse problem. The (statistical) priors on both $\mathcal{B}$ and $\mathcal{R}$ should be considered to regularize the decomposition process. By introducing priors on both $\mathcal{B}$ and $\mathcal{R}$, Ref. \cite{Jiang} proposed a tensor-based variational model for video rain streaks removal, i.e.,
%
\small
\begin{align}\label{Eq:jiang}
&\min_{\mathcal{B}, \mathcal{R}}~\alpha_{1} \|\nabla_{y} \mathcal{R}\|_{1} + \alpha_{2}\|\mathcal{R}\|_{1}+\alpha_{3}\|\nabla_{x}\mathcal{B}\|_{1} +\alpha_{4}\|\nabla_{t}\mathcal{B}\|_{1} + \mathrm{rank}(\mathcal{B}) \nonumber\\
&~~\mathrm{s.t.}~~~\mathcal{O} = \mathcal{B}+\mathcal{R}, ~ \mathcal{B}, \mathcal{R} \geq 0,
\end{align}
\normalsize
where $\nabla_{x}, \nabla_{y}, \nabla_{t}$ denote three unidirectional TV operators with respects of vertical, horizontal and temporal directions.

The tensor-based variational model (\ref{Eq:jiang}) is based on the assumption of vertical direction of falling raindrops. It is infrequent in realistic scenarios due to the influence of wind and other factors. Meanwhile, this variational model (\ref{Eq:jiang}) only considers the sparsity of vertical derivatives of rain streaks and uses the traditional TV regularizer $\|\nabla_{y} \mathcal{R}\|_{1}$. The L1-norm regularizer $\|\mathcal{R}\|_{1}$ is related to the assumption that the magnitudes of rain streaks are commonly sparse. Since the horizontal and temporal derivatives of video background are assumed to be sparse, TV regularizers $\|\nabla_{x}\mathcal{B}\|_{1}$ and $\|\nabla_{t}\mathcal{B}\|_{1}$ are used to regularize the deraining process. By using the spatio-temporal coherences of video data, tensor modeling $\mathrm{rank}(\mathcal{B})$ is introduced to further enhance imaging performance.
%
%
%
\section{Directional Regularized Tensor Modeling}
\label{sec:model}
\subsection{DTV Regularized Tensor Decomposition Model}
The falling directions of raindrops are essentially spatially variant in realistic videos. For the sake of simplicity, we assume that all rain streaks fall in an approximately identical direction. We propose to develop a directional regularized tensor modeling for removing video rain streaks as follows
\begin{align}\label{Eq:DTV}
&\min_{\mathcal{B}, \mathcal{R}}~\alpha_{1} \|\nabla_{d}\mathcal{R}\|_{1} + \alpha_{2}\|\mathcal{R}\|_{1} + \alpha_{3} \|\nabla_{x}\mathcal{B}\|_{1} \nonumber\\
&~~~~~~~~~~~~~~~~~~~~~~~~~~~~+\alpha_{4} \|\nabla_{y}\mathcal{B}\|_{1} + \alpha_{5} \|\nabla_{t}\mathcal{B}\|_{1} + \mathrm{rank} (\mathcal{B}) \nonumber\\
&~~\mathrm{s.t.}~~~\mathcal{O} = \mathcal{B} + \mathcal{R}, ~ \mathcal{B}, \mathcal{R} \geq 0,
\end{align}
where $d$ denotes the falling direction of rain streaks, $\nabla_{d} $ is the directional total variation (DTV) operator. We will detailedly discuss the potential reasons why introducing different regularization terms in Eq. (\ref{Eq:DTV}).

\textbf{Sparsity of Rain Streaks in Spatial and Derivative Domains} In the case of light rain, rain streaks can be sparsely represented in both spatial and derivative domains. As the rainfall intensity becomes higher, it is better to consider the sparsity of rain streaks in derivative domain. The image derivatives are calculated along the falling (\textit{not vertical}) direction of rain streaks. In contrast, Jiang \textit{et al.} \cite{Jiang} only considers the vertical direction which is an invalid assumption in practice. The L1-norm regularizer $\| \mathcal{R} \|_{1}$ and DTV regularizer $\| \nabla_{d}\mathcal{R} \|_{1}$ are adopted to model the sparsity of rain streaks in spatial and derivative domains, respectively.

\textbf{Sparsity of Video Intensities in Spatial and Temporal Domains} Both vertical and horizontal derivatives in outdoor images can be well distributed as the Laplacian distribution\footnote{From a statistical point of view, magnitudes of most derivatives tend to zero and few have large values for both vertical and horizontal directions.}. TV regularizer is essentially related to the assumption of Laplacian prior \cite{LiuShiMRI2014}. There commonly exists high similarity between consecutive frames within one video. The distribution of differences between any two consecutive frames (i.e., along temporal direction) can also be modeled by a Laplacian prior. Therefore, TV regularizers $\|\nabla_{x}\mathcal{B}\|_{1}$, $\|\nabla_{y}\mathcal{B}\|_{1}$ and $\|\nabla_{t}\mathcal{B}\|_{1}$ are used to model the sparsity of video intensities in vertical, horizontal and temporal directions, respectively. 

\textbf{Tensor Structures in Video Intensities} Spatial self-similarities within each frame and temporal similarities between any two consecutive frames can be visually observed, especially for video generated by static camera or with static background. Recently, low-rank \cite{Kim,Chen,RenTianCVPR2017} and tensor \cite{Jiang} modeling have significantly improved the deraining performance. By exploiting the tensor structures in video intensities \cite{Jiang}, our model (\ref{Eq:DTV}) also minimizes the rank of $ \mathcal{B}$ to promote the decomposition of video background and  rain streaks.

\subsection{Numerical Optimization}
The alternating direction method of multipliers (ADMM) framework \cite{BoydParikh2011} will be used to effectively solve the complicated unconstrained optimization problem (\ref{Eq:DTV}). By introducing several intermediate variables, we can reformulate (\ref{Eq:DTV}) as the following equivalent constrained version
\begin{align}\label{Eq:DTVconstrained}
&\min_{\mathcal{R}, \mathcal{D},\mathcal{S},\mathcal{X},\mathcal{Y},\mathcal{T},\mathcal{L}}~\alpha_{1}\| \nabla_{d}\mathcal{D}\|_{1}+\alpha_{2}\|\mathcal{S}\|_{1} +\alpha_{3}\|\mathcal{X} \|_{1}  \nonumber\\
&~~~~~~~~~~~~~~~~~~~~~~~~~~~~~~~~~~~~~~~~+\alpha_{4}\|\mathcal{Y} \|_{1} +\alpha_{5}\|\mathcal{T} \|_{1}+\|\mathcal{L} \|_{*} \nonumber\\
&~~~~~~~~~~\mathrm{s.t.}~~~\mathcal{D}=\mathcal{R},~\mathcal{S}=\mathcal{R},~\mathcal{X}=\nabla_{x}(\mathcal{O}-\mathcal{R}),\nonumber\\
&~~~~~~~~~~~~~~~~~~~\mathcal{Y}= \nabla_{y} (\mathcal{O}-\mathcal{R}),~\mathcal{T} = \nabla_{t} (\mathcal{O}-\mathcal{R}),\nonumber\\
&~~~~~~~~~~~~~~~~~~~\mathcal{L} = \mathcal{O} - \mathcal{R},~\mathcal{O} \geq \mathcal{R} \geq 0
\end{align}
whose augmented Lagrangian function $\mathcal{L}_{\mathcal{A}}$ is given by
\small
\begin{align*}
& \mathcal{L}_{\mathcal{A}} = \alpha_{1}\|\nabla_{d}\mathcal{D}\|_{1}+\alpha_{2}\|\mathcal{S}\|_{1} +\alpha_{3}\|\mathcal{X}\|_{1}+\alpha_{4}\|\mathcal{Y}\|_{1}+\alpha_{5}\|\mathcal{T}\|_{1}\\
& + \|\mathcal{L}\|_{*} +\frac{\beta_{1}}{2} \Big\| \mathcal{D} - \mathcal{R} - \frac{\Lambda_{1}}{\beta_{1}} \Big\|_{F}^{2} + \frac{\beta_{2}}{2} \Big\| \mathcal{S} - \mathcal{R} - \frac{\Lambda_{2}}{\beta_{2}} \Big\|_{F}^{2} \\
& + \frac{\beta_{3}}{2} \Big\|  \mathcal{X} - \nabla_{x}(\mathcal{O} - \mathcal{R}) - \frac{\Lambda_{3}}{\beta_{3}} \Big\|_{F}^{2} + \frac{\beta_{4}}{2} \Big\|  \mathcal{Y} - \nabla_{y}(\mathcal{O} - \mathcal{R}) - \frac{\Lambda_{4}}{\beta_{4}} \Big\|_{F}^{2}\\
& + \frac{\beta_{5}}{2} \Big\|  \mathcal{T} - \nabla_{t}(\mathcal{O} - \mathcal{R}) - \frac{\Lambda_{5}}{\beta_{5}} \Big\|_{F}^{2} + \frac{\beta_{6}}{2} \Big\|  \mathcal{L} - (\mathcal{O} - \mathcal{R}) - \frac{\Lambda_{6}}{\beta_{6}} \Big\|_{F}^{2}
\end{align*}
\normalsize
where $\Lambda = [\Lambda_{1}, \Lambda_{2}, \cdots, \Lambda_{6}]$ are Lagrangian multipliers and $\beta=[\beta_{1}, \beta_{2}, \cdots, \beta_{6}]$ are positive parameters. Within our optimization framework, ADMM minimizes $\mathcal{L}_{\mathcal{A}}$ over $\mathcal{R}$, $\mathcal{D}$, $\mathcal{S}$, $\mathcal{X}$, $\mathcal{Y}$, $\mathcal{T}$ and $\mathcal{L}$ leading to subproblems whose solutions could be efficiently obtained using simple numerical methods.

\textbf{$\mathcal{D}$-subproblem:}\quad Within the ADMM framework, the $\mathcal{D}$-subproblem can be formulated as follows
\begin{equation}\label{eq:subproblemD}
   \min_{D} \alpha_{1}\|\nabla_{d}\mathcal{D}\|_{1}+\frac{\beta_{1}}{2}\|\mathcal{R}-\mathcal{D} + \frac{\Lambda_{1}}{\beta_{1}}\|_{F}^{2}.
\end{equation}

The falling direction $d$ of rain streaks is pre-estimated directly using the Fourier transform method \cite{DTV}. The generalized proximal operator introduced in \cite{BayramSPL2012} is adopted to effectively handle the nonsmooth optimization problem (\ref{eq:subproblemD}).

\textbf{$\left( \mathcal{S},\mathcal{X},\mathcal{Y},\mathcal{T} \right)$-subproblems:} For $\mathcal{S}$-subproblem, the minimization problem can be formulated as follows

\begin{equation}\label{eq:subproblemS}
\min_{S} \alpha_{2}\|\mathcal{S}\|_{1}+\frac{\beta_{2}}{2}\|\mathcal{R}-\mathcal{S}+\frac{\Lambda_{2}}{\beta_{2}}\|_{F}^{2},
\end{equation}
which has a closed-form solution through the following soft-thresholding operator

\begin{equation}\label{eq:subproblemSsolution}
\mathcal{S}^{(t+1)} = \mathrm{Soft}_{{\alpha_{2}}/{\beta_{2}}}\left( \mathcal{R}^{(t)} + {\Lambda_{2}^{(t)}}/{\beta_{2}}\right),
\end{equation}
with $\mathrm{Soft}_{{\alpha}/{\beta}}(\cdot)$ being defined as $\mathrm{Soft}_{{\alpha}/{\beta}} \left( x \right) = \mathrm{sign}(x) \circ \max(\left| x \right|-{\alpha}/{\beta},0)$. Analogous to the $\mathcal{S}$-subproblem (\ref{eq:subproblemS}), the solutions $ \mathcal{X},\mathcal{Y},\mathcal{T}$ can be updated as follows
\begin{align}
   \mathcal{X}^{(t+1)} &= \mathrm{Soft}_{{\alpha_{3}}/{\beta_{3}}}\left(\nabla_{x} (\mathcal{O}-\mathcal{R}^{(t)})+{\Lambda_{3}^{(t)}}/{\beta_{3}} \right), \label{eq:subproblemX}\\
   \mathcal{Y}^{(t+1)} &= \mathrm{Soft}_{{\alpha_{4}}/{\beta_{4}}}\left(\nabla_{y} (\mathcal{O}-\mathcal{R}^{(t)})+{\Lambda_{4}^{(t)}}/{\beta_{4}} \right), \label{eq:subproblemY}\\
   \mathcal{T}^{(t+1)} &= \mathrm{Soft}_{{\alpha_{5}}/{\beta_{5}}}\left(\nabla_{t} (\mathcal{O}-\mathcal{R}^{(t)})+{\Lambda_{5}^{(t)}}/{\beta_{5}} \right). \label{eq:subproblemT}
\end{align}

\textbf{$\mathcal{L}$-subproblem:} The $\mathcal{L}$-subproblem is given by

\begin{equation}\label{eq:subproblemL}
   \min_{\mathcal{L}} \|\mathcal{L}\|_{*}+\frac{\beta_{6}}{2}\|\mathcal{O}-\mathcal{R}-\mathcal{L}+\frac{\Lambda_{6}}{\beta_{6}}\|_{F}^{2},
\end{equation}
where the tensor nuclear norm is defined as $\left\| \mathcal{X} \right\|_{*} = \sum_{i=1}^{n} \left\| \mathbf{X}_{i} \right\|_{*}$ with $ \mathbf{X}_{i} = \mathrm{unfold}_{i} (\mathcal{X})$. The solution $\mathcal{L}$ of (\ref{eq:subproblemL}) is given by
\begin{equation}\label{eq:subproblemLsolution}
   \mathcal{L}^{(t+1)} = \sum\nolimits_{i=1}^{3} \frac{1}{3} \mathrm{Fold}_{i} \big( \mathbf{L}_{(i)}^{(t+1)} \big),
\end{equation}
where $\mathbf{L}_{(i)}^{(t+1)}=\mathcal{Q}_{\frac{1}{\beta_{6}}}\left(\mathbf{B}_{(i)}^{(t)}+{\Lambda_{(i)}^{(t)}}/{\beta_{6}}\right)$
and $\mathcal{Q}_{\frac{1}{\beta_{6}}}(\mathbf{X})$ denotes the soft-thresholding to the singular values of $\mathbf{X}$.

\textbf{$\mathcal{R}$-subproblem:} Finally, the $\mathcal{R}$-subproblem is given by
\small
\begin{align}
   & \min_{\mathcal{R}} ~ \frac{\beta_{1}}{2}\|\mathcal{R}-\mathcal{D}\|_{F}^{2}+\frac{\beta_{2}}{2}\|\mathcal{R}-\mathcal{S}\|_{F}^{2}\\
   & + \frac{\beta_{3}}{2} \Big\|  \mathcal{X} - \nabla_{x}(\mathcal{O} - \mathcal{R}) - \frac{\Lambda_{3}}{\beta_{3}} \Big\|_{F}^{2} + \frac{\beta_{4}}{2} \Big\|  \mathcal{Y} - \nabla_{y}(\mathcal{O} - \mathcal{R}) - \frac{\Lambda_{4}}{\beta_{4}} \Big\|_{F}^{2} \nonumber\\
   & + \frac{\beta_{5}}{2} \Big\|  \mathcal{T} - \nabla_{t}(\mathcal{O} - \mathcal{R}) - \frac{\Lambda_{5}}{\beta_{5}} \Big\|_{F}^{2} + \frac{\beta_{6}}{2} \Big\|  \mathcal{L} - (\mathcal{O} - \mathcal{R}) - \frac{\Lambda_{6}}{\beta_{6}} \Big\|_{F}^{2}, \nonumber
\end{align}
\normalsize
which has the following closed-form solution
\begin{equation}\label{eq:subproblemR}
   \mathcal{R}^{(t+1)}=\mathcal{F}^{-1}\left( {\mathcal{F}(\mathcal{K}_{1})}/{\mathcal{F}(\mathcal{K}_{2})} \right),
\end{equation}
where $\mathcal{F}$ and $\mathcal{F}^{-1}$ denote the fast Fourier transform (FFT) and its inverse, $\mathcal{K}_{1}=\beta_{1}\mathcal{D}^{(t+1)}-\Lambda_{1}^{t}
+\beta_{2}\mathcal{S}^{(t+1)}-\Lambda_{2}^{(t)} +\nabla_{x}^{T}(\beta_{3}\nabla_{x}\mathcal{O}-\beta_{3}\mathcal{X}^{(t+1)}+\Lambda_{3}^{(t)})+ \nabla_{y}^{T}(\beta_{4}\nabla_{y}\mathcal{O}-\beta_{4}\mathcal{Y}^{(t+1)}+\Lambda_{4}^{(t)}) + \nabla_{t}^{T}(\beta_{5}\nabla_{t}\mathcal{O}-\beta_{5}\mathcal{T}^{(t+1)}+\Lambda_{5}^{(t)}) +\beta_{6}(\mathcal{O}-\mathcal{L}^{(t+1)})+\Lambda_{6}^{(t)}$ and $\mathcal{K}_{2}= (1 + \beta_{1}+\beta_{2}) \mathcal{I}+\beta_{3}\nabla_{x}^{T}\nabla_{x} +\beta_{4}\nabla_{y}^{T}\nabla_{y}+\beta_{5}\nabla_{t}^{T}\nabla_{t}$.
%

\textbf{$\Lambda$ update:} During each iteration, the Lagrangian multipliers $\Lambda$ are updated using $\Lambda_{1}^{(t+1)} = \Lambda_{1}^{(t)} + \beta_{1} (\mathcal{R}^{(t+1)}-\mathcal{D}^{(t+1)})$, $\Lambda_{2}^{(t+1)} = \Lambda_{2}^{(t)} + \beta_{2} (\mathcal{R}^{(t+1)}-\mathcal{S}^{(t+1)})$, $\Lambda_{3}^{(t+1)} = \Lambda_{3}^{(t)} + \beta_{3} (\nabla_{x}(\mathcal{O}-\mathcal{R}^{(t+1)})-\mathcal{X}^{(t+1)})$, $\Lambda_{4}^{(t+1)} = \Lambda_{4}^{(t)} + \beta_{4} (\nabla_{y}(\mathcal{O}-\mathcal{R}^{(t+1)})-\mathcal{Y}^{(t+1)})$, $\Lambda_{5}^{(t+1)} = \Lambda_{5}^{(t)} + \beta_{5} (\nabla_{t}(\mathcal{O}-\mathcal{R}^{(t+1)})-\mathcal{T}^{(t+1)})$ and $\Lambda_{6}^{(t+1)} = \Lambda_{6}^{(t)} + \beta_{6} (\mathcal{O}-\mathcal{R}^{(t+1)}-\mathcal{L}^{(t+1)})$.
%
%
%
\setlength{\tabcolsep}{1.2pt}
\begin{table*}[t]
	\scriptsize
	\caption{Quantitative comparisons of rain streaks removal on five different synthetic videos.}
		\begin{tabular}{l|l|cccc|cccc|cccc|cccc}
			\hline
			\multicolumn{2}{c}{\multirow{2}{*}{Rain Types}} & \multicolumn{8}{c}{Heavy} & \multicolumn{8}{c}{Light}\\
			\cline{3-18}
			\multicolumn{2}{c}{}  & \multicolumn{4}{c}{45 Degrees} & \multicolumn{4}{c}{60 Degrees}& \multicolumn{4}{c}{45 Degrees}& \multicolumn{4}{c}{60 Degrees}\\
			\hline
			\multicolumn{2}{c}{}
			&PSNR($\mathcal{B}$)&SSIM($\mathcal{B}$)&SSIM($\mathcal{R}$)&RES($\mathcal{B}$)
			&PSNR($\mathcal{B}$)&SSIM($\mathcal{B}$)&SSIM($\mathcal{R}$)&RES($\mathcal{B}$)
			&PSNR($\mathcal{B}$)&SSIM($\mathcal{B}$)&SSIM($\mathcal{R}$)&RES($\mathcal{B}$)
			&PSNR($\mathcal{B}$)&SSIM($\mathcal{B}$)&SSIM($\mathcal{R}$)&RES($\mathcal{B}$)\\
			\hline
			\multirow{5}{*}{\rotatebox{90}{Foreman}}
			&Rainy      &27.519     &0.719      &-           &33.211     &27.537     &0.721      &-         &33.150
			&36.279     &0.828      &-           &12.126     &36.632     &0.832      &-         &11.650\\
			&Luo \textit{et al.} \cite{Luo} &29.693     &0.721      &\textbf{0.685}     &25.858     &30.006     &0.728  &\textbf{0.631}     &24.959
			&35.115     &0.822      &0.251       &13.868     &34.996     &0.827      &0.273     &14.058\\
			&Li \textit{et al.} \cite{Li}  &31.773     &0.771      &0.382       &20.361     &31.827     &0.771      &0.377     &\textbf{20.237}
			&34.914     &0.838      &0.230       &14.187     &34.911     &0.836      &0.238     &14.190\\
			&Jiang \textit{et al.} \cite{Jiang} &30.008   &0.788      &0.316       &24.996     &30.57	     &0.813      &0.383     &23.408	
			&38.009     &0.876      &0.438       &10.355	 &38.970     &0.903      &0.533     &9.460\\
			&Ours   &\textbf{31.853}	   &\textbf{0.852}	    &0.536	         &\textbf{20.226}
			&\textbf{31.841}       &\textbf{0.852}      & 0.531          &20.269	
			&\textbf{39.401}       &\textbf{0.947}	    &\textbf{0.669}	 &\textbf{9.399}
			&\textbf{39.388}       &\textbf{0.941}	    &\textbf{0.645}	 &\textbf{9.426}
			\\
			\hline
			\multirow{5}{*}{\rotatebox{90}{Girl}}
			&Rainy          &28.596     &0.732      &-                  &48.881     &28.588     &0.731      &-                   &48.920
			&37.348     &0.867      &-                  &17.859     &38.141     &0.867	    &-                   &19.965\\
			&Luo\textit{et al.} \cite{Luo}     &30.631     &0.745      &\textbf{0.821}	    &38.670	    &30.534	    &0.743      &\textbf{0.831}      &39.119
			&37.154	    &0.865      &0.116              &18.265     &37.088     &0.867	    &0.013               &18.397\\
			&Li \textit{et al.} \cite{Li}      &32.411     &0.719      &0.309              &31.507	    &32.474     &0.722      &0.301	             &31.278	
			&34.468	    &0.775      &0.229              &24.854     &34.479     &0.777      &0.230               &24.829\\
			&Jiang \textit{et al.} \cite{Jiang}   &31.392	    &0.834      &0.346              &35.828     &32.018     &0.852      &0.418               &33.445
			&\textbf{38.809}&0.936  &0.636              &\textbf{17.082}     &\textbf{41.094}&0.940  &0.718               &\textbf{16.504}\\
			&Ours       &\textbf{33.083}    &\textbf{0.882}	    &0.559          &\textbf{29.936}
			&\textbf{32.906}    &\textbf{0.881}	    &0.560          &\textbf{30.522}
			&38.601             &\textbf{0.976}     &\textbf{0.823}  &18.137
			&38.016             &\textbf{0.975}	    &\textbf{0.817}  &19.140
			\\
			\hline
			\multirow{5}{*}{\rotatebox{90}{Highway}}
			&Rainy          &28.178     &0.530	     &-      &30.790        &28.154        &0.527	        &-      &30.883
			&36.946     &0.712	     &-      &11.245        &36.813        &0.708	        &-      &11.406\\
			&Luo \textit{et al.} \cite{Luo}     &30.564	    &0.549       &\textbf{0.735}       &23.404        &31.06	        &0.559      &\textbf{0.638}	       &22.109	
			&37.589	    &0.712       &0.246                &10.435        &37.541           &0.714	    &0.106	               &10.482\\
			&Li \textit{et al.} \cite{Li}      &30.755     &0.543       &0.390	               &22.883	         &30.741	    &0.543	     &0.390       &22.926	
			&32.926	    &0.674	     &0.108                &17.815           &32.998        &0.677	     &0.107       &17.670\\
			&Jiang \textit{et al.} \cite{Jiang}   &30.839	    &0.647       &0.325	               &22.711          &31.362         &0.673          &0.390      &21.394
			&40.154	    &0.797       &0.540	               &8.096	        &41.292         &0.828          &0.632	    &7.204\\
			&Ours       &\textbf{33.047}	    &\textbf{0.723}         &0.544	        &\textbf{17.638}
			&\textbf{33.486}        &\textbf{0.720}	        &0.615          &\textbf{16.868}	
			&\textbf{44.999}	    &\textbf{0.901}         &\textbf{0.788} &\textbf{5.057}
			&\textbf{44.711}        &\textbf{0.893}         &\textbf{0.772}	&\textbf{5.204}\\
			\hline
			\multirow{5}{*}{\rotatebox{90}{Truck}}
			&Rainy          &27.329	    &0.686       &-      &56.552         &27.345     &0.687		         &-      &56.452
			&36.303	    &0.824	     &-      &20.151         &36.204     &0.821	             &-      &20.361\\
			&Luo \textit{et al.} \cite{Luo}     &\textbf{30.882}	&0.700      &0.529       &\textbf{37.582}        &\textbf{31.962}	    &0.720       &0.327       &\textbf{33.216}	
			&33.332	            &0.813      &0.338	     &28.465	             &32.644                &0.814	     &0.409	      &30.701\\
			&Li \textit{et al.} \cite{Li}      &30.842	    &0.704       &0.297      &37.739     &30.866     &0.706       &0.303	     &37.639	
			&31.831	    &0.784       &0.136	     &33.674     &31.821     &0.784	      &0.137	     &33.715\\
			&Jiang \textit{et al.} \cite{Jiang}   &28.919     &0.749	     &0.174	     &47.213	 &29.470	 &0.775       &0.245	     &44.338	
			&38.656	    &0.876	     &0.398	     &15.711     &39.852     &0.899       &0.548	     &13.816\\
			&Ours       &30.138          &\textbf{0.840}	                &\textbf{0.620}	                &41.164
			&30.514	         &\textbf{0.852}                    &\textbf{0.628}                 &39.438
			&\textbf{45.141} &\textbf{0.968}                    &\textbf{0.832}	                &\textbf{8.194}
			&\textbf{44.141} &\textbf{0.959}	                &\textbf{0.807}	                &\textbf{9.003}
			\\
			\hline
			\multirow{5}{*}{\rotatebox{90}{Waterfall}}
			&Rainy          &27.688	            &0.804           &-          &32.577	             &27.709         &0.804	         &-          &32.503
			&36.445             &0.914	         &-          &11.910                 &36.783         &0.920	         &-          &11.451\\
			&Luo \textit{et al.} \cite{Luo}     &30.649	            &0.796           &0.349      &23.172                 &30.848         &0.798          &0.283      &22.653
			&30.454             &0.893           &0.527	     &21.986                 &31.114         &0.900	         &0.524	     &21.986\\
			&Li \textit{et al.} \cite{Li}      &31.149             &0.689           &0.188      &21.862                 &31.134	     &0.691          &0.194	     &21.898	
			&31.085	            &0.737	         &0.127      &22.030	             &30.991	     &0.736	         &0.124	     &22.264\\
			&Jiang \textit{et al.} \cite{Jiang}   &29.346	            &0.858           &0.181	     &26.959                 &29.848	     &0.877          &0.241	     &25.448	
			&38.332	            &0.938           &0.347      &9.783	                 &39.458         &0.952	         &0.459      &8.689\\
			&Ours       &\textbf{32.510}	&\textbf{0.919}     &\textbf{0.581}	    &\textbf{18.766}
			&\textbf{32.568}    &\textbf{0.920}	    &\textbf{0.581}     &\textbf{18.643}	
			&\textbf{41.531}    &\textbf{0.975}	    &\textbf{0.699}	    &\textbf{6.960}
			&\textbf{41.456}    &\textbf{0.974} 	&\textbf{0.670}	    &\textbf{7.020}
			\\
			\hline
		\end{tabular}\label{Table1}
\end{table*}
\begin{figure}[t]
	\centering
	\includegraphics[width=0.96\linewidth]{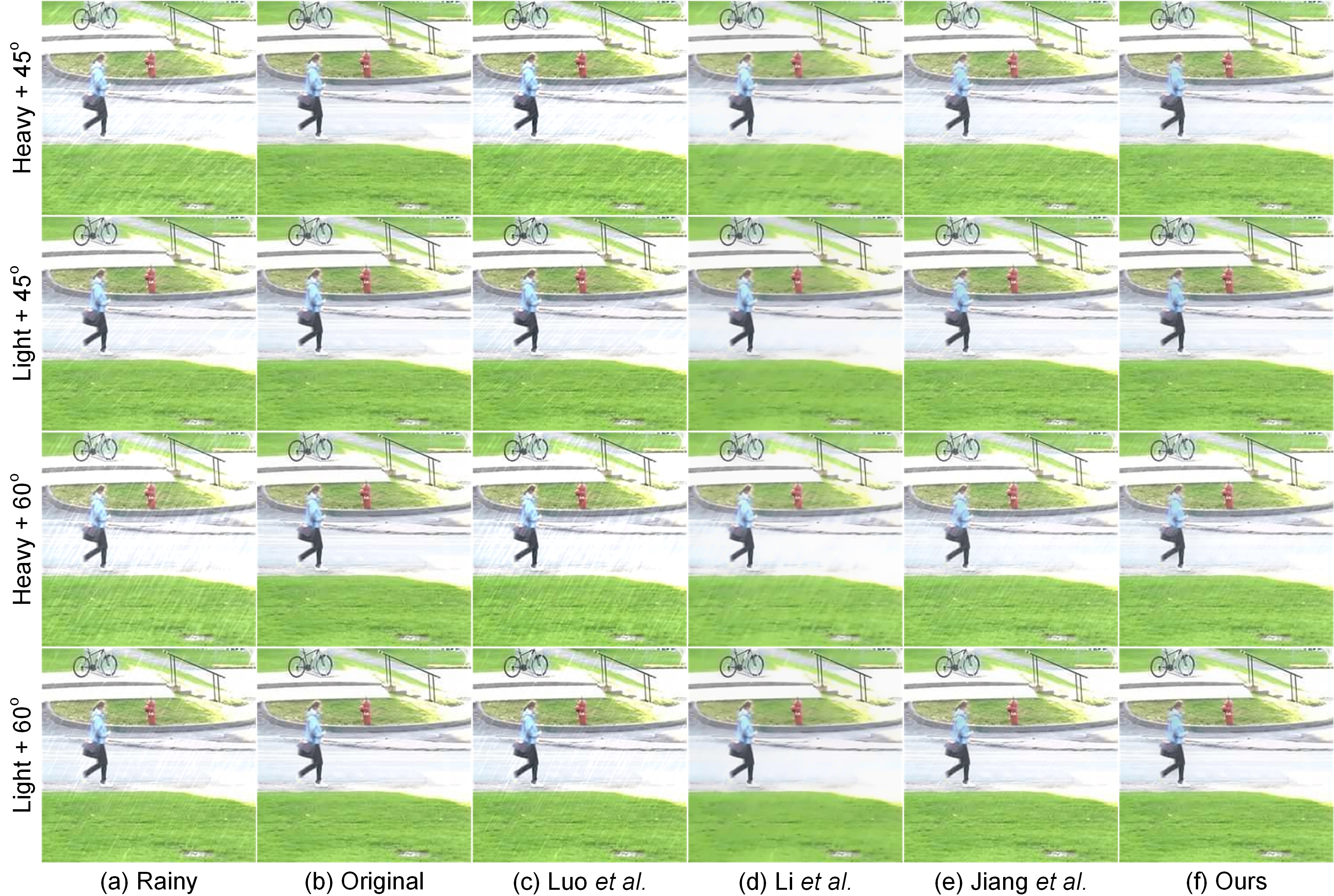}
	\caption{Comparisons of deraining results on one frame extracted from the synthetic ``Girl" video under different rainy conditions. From left to right: (a) rainy images, (b) original rain-free images, restored images generated by (c) Luo \textit{et al.} \cite{Luo}, (d) Li \textit{et al.} \cite{Li}, (e) Jiang \textit{et al.} \cite{Jiang} and (f) ours.}
	\label{Figure1} 
\end{figure}
\begin{figure}[t]
	\centering
	\includegraphics[width=0.97\linewidth]{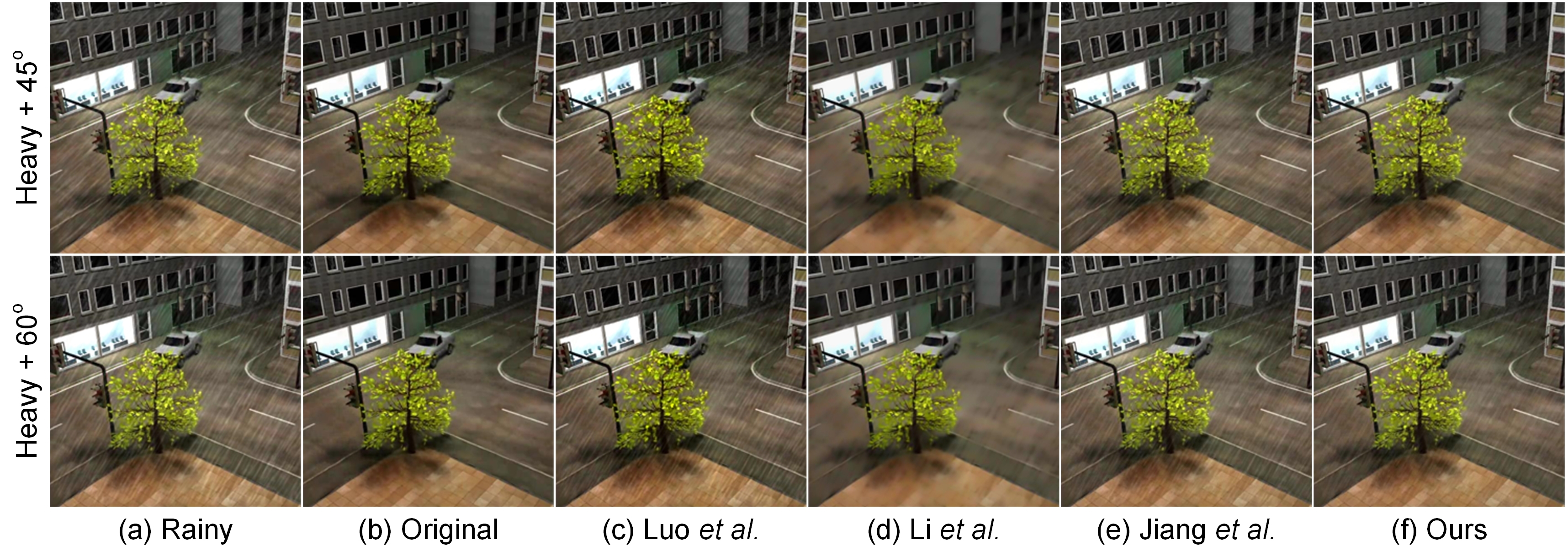}
	\caption{Comparisons of deraining results on one frame extracted from the synthetic ``Truck" video under different rainy conditions. From left to right: (a) rainy images, (b) original rain-free images, restored images generated by (c) Luo \textit{et al.} \cite{Luo}, (d) Li \textit{et al.} \cite{Li}, (e) Jiang \textit{et al.} \cite{Jiang} and (f) ours.}
	\label{Figure2}
\end{figure}
\section{Experimental Results and Discussion}
\label{sec:pagestyle}
To verify the effectiveness of the proposed method, we compare our method with several advanced deraining methods, including sparse coding method Luo \textit{et al.} \cite{Luo}, layer priors-based method Li \textit{et al.} \cite{Li}, and tensor-based method Jiang \textit{et al.} \cite{Jiang}. For synthetic experiments, the falling directions of rain streaks were randomly selected between $[40~50]$ or $[55~65]$. In numerical experiments, the regularization parameters $\{ \alpha_{1}, \alpha_{2}, \cdots, \alpha_{5} \}$ were manually selected from $\{10^{1},10^{2},10^{3} \}$ and $\{\beta_{1}, \beta_{2}, \cdots, \beta_{6}\}$ were set to be $50$. The effectiveness of these manually selected parameters has been illustrated through comprehensive deraining experiments. Our Matlab source code is available at \url{https://github.com/Snowfallingplum/rain-streaks-removal}.
\subsection{Synthetic Experiments}
%
The rain streaks with different types are manually generated and incorporated into the original video backgrounds. Objective image quality metrics, i.e., PSNR, SSIM and residual error (RES), are selected to quantitatively evaluate the deraining results. Table \ref{Table1} detailedly depicts the quantitative results for five synthetic datasets. It is shown that our method generates the best evaluation results under consideration in most of the cases. Its superior performance is further confirmed in Figs. \ref{Figure1} and \ref{Figure2}. The degrees $45^{\circ}$ and $60^{\circ}$ are the average falling directions of rain streaks. The restored images generated by Luo \textit{et al.} \cite{Luo} still suffer from rain streaks remaining leading to visual quality degradation. Li \textit{et al.} \cite{Li} could effectively reduce rain effects but tend to oversmooth fine details. Compared with Jiang \textit{et al.} \cite{Jiang}, our method is able to effectively remove more rain streaks while preserving fine details.
\subsection{Realistic Experiments}
\begin{figure}[t]
	\centering
	\includegraphics[width=0.95\linewidth]{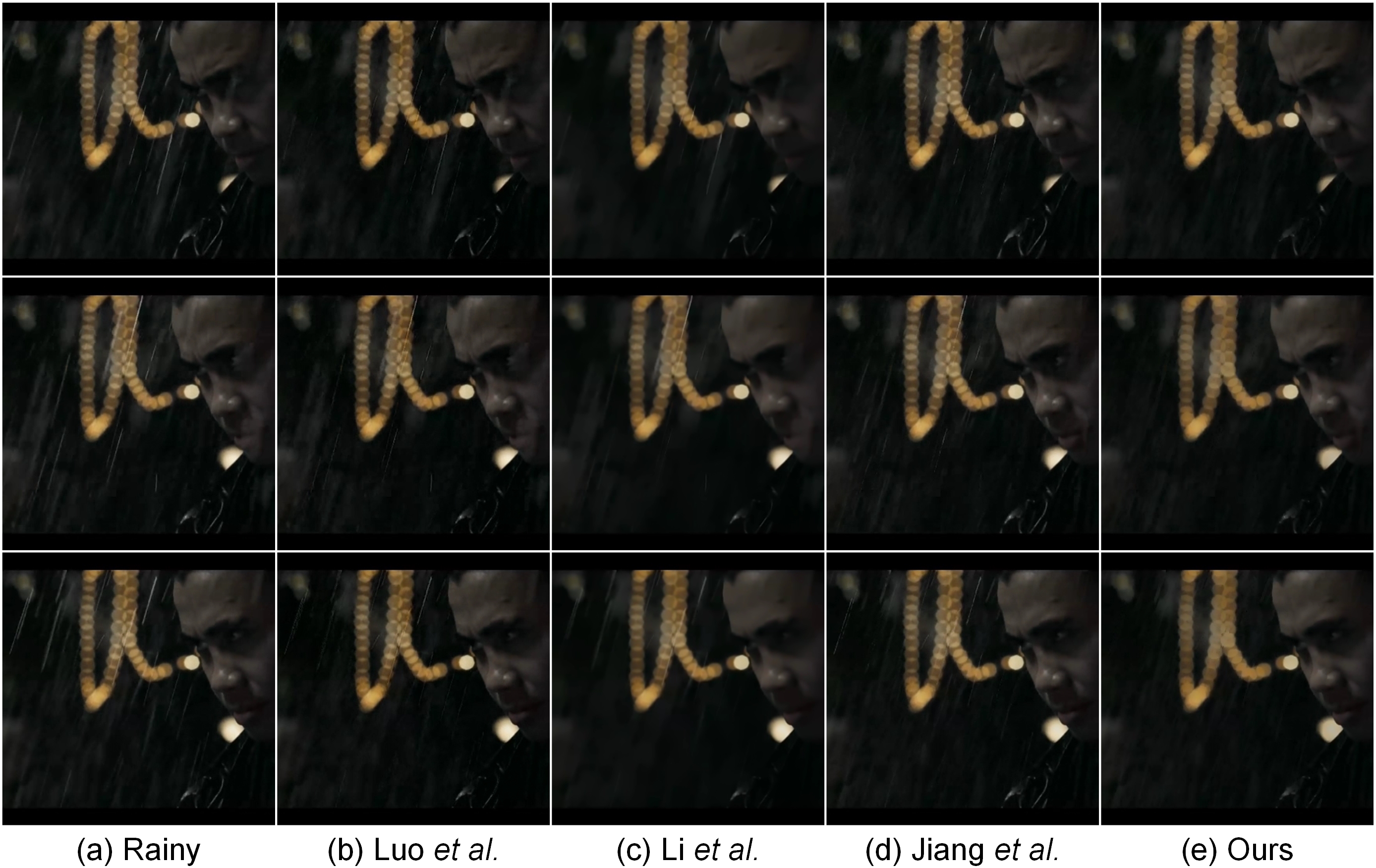}
	\caption{Comparisons of deraining results on three frames extracted from the realistic ``The Grandmaster" video. From left to right: (a) rainy images, restored images yielded by (b) Luo \textit{et al.} \cite{Luo}, (c) Li \textit{et al.} \cite{Li}, (d) Jiang \textit{et al.} \cite{Jiang} and (e) ours.}
	\label{Figure3}
\end{figure}

In the realistic experiments, it is impossible to accurately obtain the video backgrounds for quantitative evaluation. In theory, deraining methods should preserve fine details while sufficiently removing rain streaks. Fig. \ref{Figure3} illustrates the deraining results on three different frames extracted from the realistic ``The Grandmaster" video. Due to the nonvertical direction of falling raindrops, other three competing methods fail to effectively remove the undesirable rain streaks. In contrast, the proposed method fits this realistic scene well and removes almost all rain streaks. Its superior performance benefits from the DTV regularizer and tensor modeling.
\section{Conclusion}
\label{sec:conclusion}
We proposed a directional regularized tensor modeling for video rain streaks removal in this work. The main contribution is the assumption that all rain streaks fall in an approximately similar arbitrary direction. Experiments show that the proposed method can effectively remove rain streaks while preserving fine details. It is worth pointing out that our method suffers mainly from two limitations: (1) high computational complexity; (2) difficultly handling extremely heavy rain streaks. These issues are the focuses of our future work.
%
%


\begin{thebibliography}{0}
	%
    \bibitem{de1}
    M. Roser and A. Geiger, ``Video-based raindrop detection for improved image registration," in \emph{Proc. IEEE ICCV Workshops}, Kyoto, Japan, Sep. 2009, pp. 570-577. 
    
    \bibitem{de2}
    S. Maji, A. C. Berg, and J. Malik, ``Classification using intersection kernel support vector machines is efficient," in \emph{Proc. IEEE CVPR}, Anchorage, AK, USA,Jun. 2008, pp. 1-8.

	\bibitem{event}
	M. S. Shehata, J. Cai, W. M. Badawy, T. W. Burr, M. S. Pervez, R. J. Johannesson, and A. Radmanesh, ``Video-based automatic incident detection for smart roads: The outdoor environmental challenges regarding false alarms," \emph{IEEE Trans. Intell. Transp. Syst.}, vol. 9, no. 2, pp. 349-359, Jun. 2008.

	\bibitem{recognition}
	K. Garg and S. K. Nayar, ``Vision and rain," \emph{Int. J. Comput. Vision}, vol. 75, no. 1, pp. 3-27, Oct. 2007

	\bibitem{track}
	D. Comaniciu, V. Ramesh, and P. Meer, ``Kernel-based object tracking," \emph{IEEE Trans. Pattern Anal. Mach. Intell.}, vol. 25, no. 5, pp. 564-577, May 2003.

	\bibitem{Garg}
	K. Garg and S. K. Nayar, ``Detection and removal of rain from videos," in \emph{Proc. IEEE CVPR}, Washington, DC, USA, Jun. 2004, pp. 1-8.

	\bibitem{Kang}
	L. W. Kang, C. W. Lin, and Y. H. Fu, ``Automatic single-image-based rain streaks removal via image decomposition," \emph{IEEE Trans. Image Process.}, vol. 21, no. 4, pp. 1742-1755, Apr. 2012.
	
	\bibitem{Luo}
	Y. Luo, Y. Xu, and H. Ji, ``Removing rain from a single image via discriminative sparse coding," in \emph{Proc. IEEE ICCV}, Santiago, Chile, Dec. 2015, pp. 3397-3405.
	
	\bibitem{Li}
	Y. Li, R. T. Tan, X. Guo, J. Lu, and M. S. Brown, ``Rain streak removal using layer priors," in \emph{Proc. IEEE CVPR}, Las Vegas, NV, USA, Jun. 2016, pp. 2736-2744.
	
	\bibitem{Kim}
	J. H. Kim, J. Y. Sim, and C. S. Kim, ``Video deraining and desnowing using temporal correlation and low-rank matrix completion," \emph{IEEE Trans. Image Process.}, vol. 24, no. 9, pp. 2658-2670, Sep. 2015.
	
	\bibitem{Jiang}
	T. X. Jiang, T. Z. Huang, X. L. Zhao, L. J. Deng, and Y. Wang, ``A novel tensor-based video rain streaks removal approach via utilizing discriminatively intrinsic priors," in \emph{Proc. IEEE CVPR}, Honolulu, HI, USA, Jul. 2017, pp. 2818-2827.
	
	\bibitem{Wei}
	W. Wei, L. Yi, Q. Xie, Q. Zhao, D. Meng, and Z. Xu, ``Should we encode rain streaks in video as deterministic or stochastic," in \emph{Proc. IEEE ICCV}, Venice, Italy, Oct. 2017, pp. 2516-2525.
	
	\bibitem{review}
	A. K. Tripathi and S. Mukhopadhyay, ``Removal of rain from videos: A review," \emph{Signal, Image and Video Processing}, vol. 8, no. 8, pp. 1421-1430, Nov. 2014.
	
	\bibitem{Yang}
	W. Yang, R. T. Tan, J. Feng, J. Liu, Z. Guo, and S. Yan, ``Deep joint rain detection and removal from a single image," in \emph{Proc. IEEE CVPR}, Honolulu, HI, USA, Jul. 2017, pp. 1357-1366.
	
	\bibitem{Fu}
	X. Fu, J. Huang, X. Ding, Y. Liao, and J. Paisley, ``Clearing the skies: A deep network architecture for single-image rain removal," \emph{IEEE Trans. Image Process.}, vol. 26, no. 6, pp. 2944-2956, Jun. 2017.
	
	\bibitem{Liu}
	J. Liu, W. Yang, S. Yang, and Z. Guo, ``Erase or fill? Deep joint recurrent rain removal and reconstruction in videos," in \emph{Proc. IEEE CVPR}, Salt Lake City, UT, USA, Jun. 2018, pp. 3233-3242.
	
	\bibitem{LiuYangTIP2019}
	J. Liu, W. Yang, S. Yang, and Z. Guo, ``D3R-Net: Dynamic routing residue recurrent network for video rain removal," \emph{IEEE Trans. Image Process.}, vol. 28, no. 2, pp. 699-712, Feb. 2019.
	
	\bibitem{LiuJiangICASSP2018}
	R. Liu, Z. Jiang, L. Ma, X. Fan, H. Li, and Z. Luo, ``Deep layer prior optimization for single image rain streaks removal," in \emph{Proc. IEEE ICASSP}, Calgary, AB, Canada, Apr. 2018, pp. 1408-1412.
	
	\bibitem{LiuShiMRI2014}
	R. W. Liu, L. Shi, W. Huang, J. Xu, S. C. H. Yu, and D. Wang, ``Generalized total variation-based MRI Rician denoising model with spatially adaptive regularization parameters," \emph{Magn. Reson. Imaging}, vol. 32, no. 6, pp. 702-720, Jun. 2014.
	
	\bibitem{Chen}
	Y. L. Chen and C. T. Hsu, ``A generalized low-rank appearance model for spatio-temporally correlated rain streaks," in \emph{Proc. IEEE ICCV}, Sydney, NSW, Australia, Dec. 2013, pp. 1968-1975.
	
	\bibitem{RenTianCVPR2017}
	W. Ren, J. Tian, Z. Han, A. Chan, and Y. Tang, ``Video desnowing and deraining based on matrix decomposition," in \emph{Proc. IEEE CVPR}, Honolulu, HI, USA, Jul. 2017, pp. 4210-4219.
	
	\bibitem{BoydParikh2011}
	S. Boyd, N. Parikh, E. Chu, B. Peleato, and J. Eckstein, ``Distributed optimization and statistical learning via the alternating direction method of multipliers," \emph{Found. Trends Mach. Learn.}, vol. 3, no. 1, pp. 1-122, 2011.
	
	\bibitem{DTV}
	R. D. Kongskov and Y. Dong, ``Directional total generalized variation regularization for impulse noise removal," in \emph{Proc. SSVM}, Graz, Austria, Jun. 2017, pp. 221-231.
	
	\bibitem{BayramSPL2012}
	\.{I}. Bayram and M. E. Kamasak, ``Directional total variation," \emph{IEEE Signal Process Lett.}, vol. 19, no. 12, pp. 781-784, Dec. 2012.
	
\end{thebibliography}
\end{document}